\documentclass[sn-mathphys-num]{sn-jnl}

\usepackage{graphicx}%
\usepackage{multirow}%
\usepackage{amsmath,amssymb,amsfonts}%
\usepackage{amsthm}%
\usepackage{mathrsfs}%
\usepackage[title]{appendix}%
\usepackage{xcolor}%
\usepackage{textcomp}%
\usepackage{manyfoot}%
\usepackage{booktabs}%
\usepackage{algorithm}%
\usepackage{algorithmicx}%
\usepackage{algpseudocode}%
\usepackage{listings}%
\def\etal{\emph{et al.}}

\theoremstyle{thmstyleone}%
\theoremstyle{thmstyletwo}%

\theoremstyle{thmstylethree}%

\begin{document}

\title[Article Title]{Planning and Rendering: Towards Product Poster Generation with Diffusion Models}


\author[1,2]{\fnm{Zhaochen} \sur{Li}}\email{zhaochenli@pku.edu.cn}
\equalcont{These authors contributed equally to this work.}

\author[1,3]{\fnm{Fengheng} \sur{Li}}\email{lifengheng@mail.nankai.edu.cn}
\equalcont{These authors contributed equally to this work.}

\author*[1]{\fnm{Wei} \sur{Feng}}\email{fengwei25@jd.com}
\equalcont{These authors contributed equally to this work.}

\author[1]{\fnm{ Honghe} \sur{Zhu}}\email{zhuhonghe1@jd.com}
\author[1]{\fnm{ Yaoyu} \sur{Li}}\email{liyaoyu1@jd.com}
\author[1]{\fnm{ Zheng} \sur{Zhang}}\email{zhangzheng11@jd.com}
\author[1]{\fnm{ Jingjing} \sur{Lv}}\email{lvjingjing1@jd.com}
\author[1]{\fnm{ Junjie} \sur{Shen}}\email{shenjunjie@jd.com}
\author[1]{\fnm{ Zhangang} \sur{Lin}}\email{linzhangang@jd.com}
\author[1]{\fnm{ Jingping} \sur{Shao}}\email{shaojingping@jd.com}
\author[3]{\fnm{ Zhenglu} \sur{Yang}}\email{yangzl@nankai.edu.cn}

\affil[1]{\orgdiv{Retail Platform Operation and Marketing Center}, \orgname{JD}, \orgaddress{ \city{Beijing}, \country{China}}}

\affil[2]{\orgname{Peking University}, \orgaddress{\city{Beijing}, \country{China}}}

\affil[3]{\orgname{Nankai University}, \orgaddress{\city{Tianjin}, \country{China}}}


\abstract{Product poster generation significantly optimizes design efficiency and reduces production costs. Prevailing methods predominantly rely on image-inpainting methods to generate clean background images for given products. Subsequently, poster layout generation methods are employed to produce corresponding layout results. However, the background images may not be suitable for accommodating textual content due to their complexity, and the fixed location of products limits the diversity of layout results. To alleviate these issues, we propose a novel product poster generation framework based on diffusion models named P\&R. The P\&R draws inspiration from the workflow of designers in creating posters, which consists of two stages: Planning and Rendering. At the planning stage, we propose a PlanNet to generate the layout of the product and other visual components considering both the appearance features of the product and semantic features of the text, which improves the diversity and rationality of the layouts. At the rendering stage, we propose a RenderNet to generate the background for the product while considering the generated layout, where a spatial fusion module is introduced to fuse the layout of different visual components. To foster the advancement of this field, we propose the first product poster generation dataset PPG30k, comprising 30k exquisite product poster images along with comprehensive image and text annotations. Our method outperforms the state-of-the-art product poster generation methods on PPG30k. The PPG30k will be released soon.}




\maketitle

\section{Introduction}
Product posters are important for product promotion. An exquisite poster should not only have a rational layout of visual components, such as underlays, texts, and the product but also have a pleasant background. This challenging task is typically undertaken by professional designers, which leads to low efficiency and high cost in poster production. To generate high-quality posters with low cost, we propose a new task, Product Poster Generation (PPG), which aims to generate an image that transmits the product messages to social groups given the product image and texts.

\begin{figure}
  \centering
  \includegraphics[width=\linewidth]{ 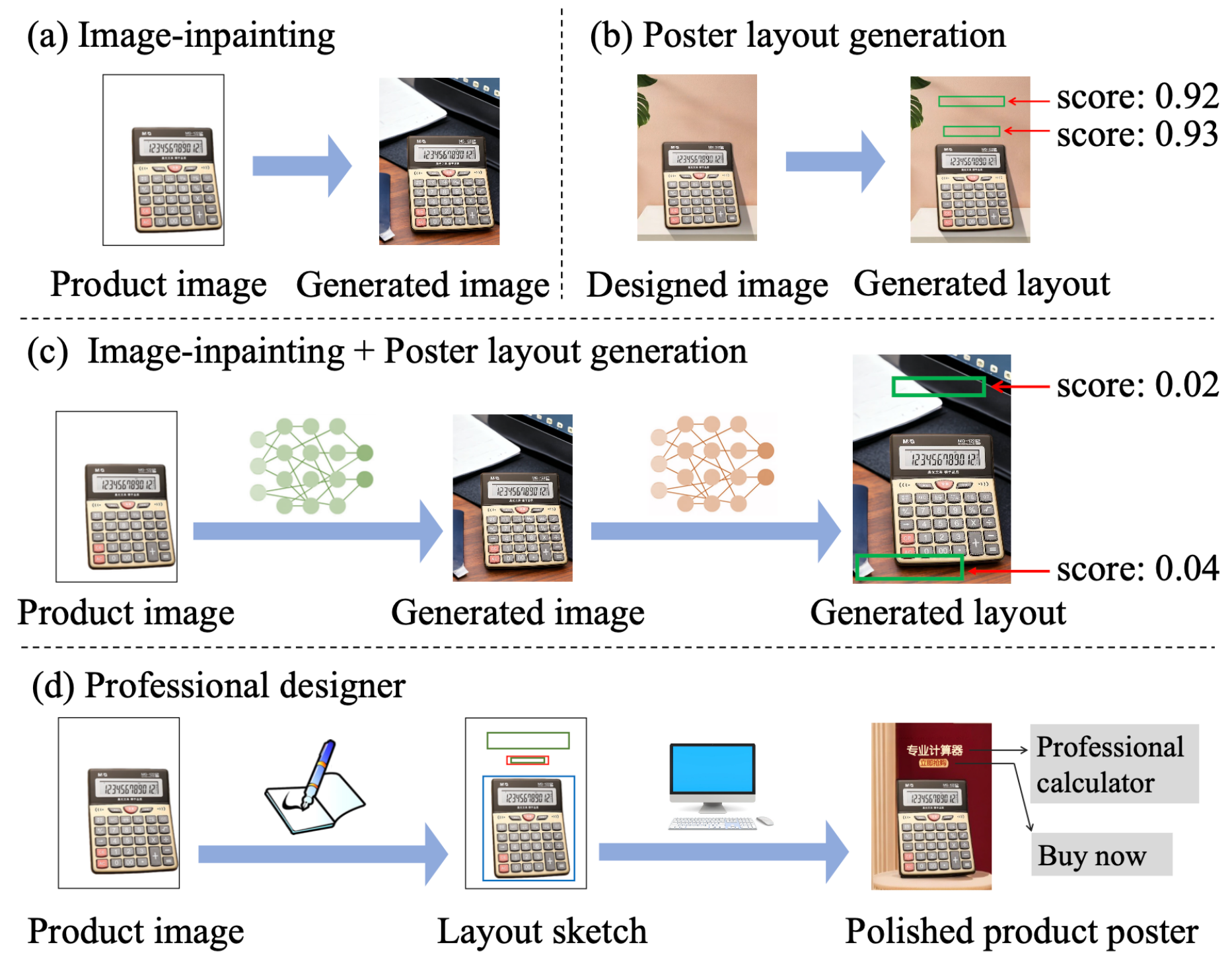}
  \caption{Two fields related to CPG: (a) Image-inpainting and (b) Poster layout generation. (c) The combination of (a) and (b) can be regarded as a naive solution to CPG. (d) The workflow of professional designers. ``score''  refers to the confidence of layout prediction, and the gray blocks are translations of texts. Green and red boxes represent texts and underlays, respectively.
}
  \label{fig1}
\end{figure}

The most related fields to PPG are image-inpainting~\cite{ku2023staging,lama,jiahui,repaint,blended,controlnet,Rombach_2022_CVPR} and poster layout generation~\cite{cgl-gan,xu2023unsupervised,Hsu-2023-posterlayout,fengheng}. As Figure~\ref{fig1} (a) and (b) show, the image-inpainting model completes the background of the poster based on the product image, and the poster layout generator plans the layout of visual components given a manually designed image. As Figure~\ref{fig1} (c) shows, simply combining the two methods can be regarded as a naive solution to PPG, where the result of the image-inpainting model can be put into the layout generator to arrange the visual components. However, this solution still has the following issues: 1) The background generated from the image-inpainting model may be too complex to display any text, which leads to a higher failure rate of layout prediction. 2) The image-inpainting methods pre-fix the location of the product, so that the layout generator can only arrange the locations of texts and underlays. This limits the diversity of poster design.  

Let's revisit how professional designers tackle the challenges of this task. As Figure~\ref{fig1} (d) shows, most designers use a hybrid process that consists of two stages~\cite{wiki}: Planning and rendering. At the planning stage, they sketch layouts of the product and other visual components to execute their ideas. The positions of other visual components get rid of the constraint of the predetermined product location, which contributes to the diversity of the layouts. At the rendering stage, they place the visual components based on the layout results and polish them on a computer. They consider the positions of texts when drawing the background, thereby avoiding that the drawn background is too complex to place texts. 

Referring to the designing process of professional designers, we put forward the first product poster generation framework P\&R with diffusion models, shown as Figure~\ref{fig: framework}. At the planning stage, we propose the PlanNet to generate layouts of products and other visual components. PlanNet firstly encodes the product image and the texts, and then fuses them with a layout decoder to predict more reasonable and flexible layouts.

At the rendering stage, we propose the RenderNet to integrate layouts and the product image into control conditions, where a spatial fusion module is used to explore the spatial relationships among layouts generated by PlanNet, and the appearance of the product is also encoded to ensure harmony between the generated background and the product. 

To train and evaluate our method, we construct the first product poster generation dataset PPG30k including appealing product posters, layouts of visual components, product masks, and text contents. We conduct comprehensive experiments on PPG30k, and the results indicate that our method performs better in terms of the rationality of the layout and the aesthetics of the image.

The contributions of our paper are four-fold:
\begin{itemize}
    \item We propose a new task PPG and the first product poster generation framework P\&R, which can generate appealing posters to transmit product messages effectively.
    \item We put forward the PlanNet to plan the overall poster layout, taking into account the product appearance and text content comprehensively.
    \item We put forward the RenderNet to render the overall layout into a poster image, where product images and layouts of visual components are integrated into multiple conditions to control the generation process.
    \item We build the first large-scale dataset PPG30k for training and evaluating the models of PPG. 
\end{itemize}

Our paper is organized as follows: In Section \ref{sec:formatting}, we summarize the latest development in layout and conditional image generation.  In Section \ref{sec:method}, we show the details of our proposed method P\&R. In Section \ref{sec:dataset}, we introduce the construction of our dataset PPG30k. In Section \ref{sec:exp}, we compare our proposed method with SOTA methods. In the last section, we draw a conclusion and discuss the future work.
\section{Related Work}
\label{sec:formatting}

\subsection{Layout Generation}
Previous layout generation methods~\cite{Cao2012, ODonovan2014LearningLF, Pang2016DirectingUA} mainly rely on heuristic algorithms and templates, which have limitations in the scope of application and lack diversity. With the advances in deep learning, a variety of deep models including GAN~\cite{gan,zheng-sig19,Li2019LayoutGANGG, cgl-gan,xu2023unsupervised}, VAE~\cite{jyothi2019layoutvae}, transformers~\cite{vaswani2017transformer,BLT2022ECCV, Gupta2020LayoutTransformerLG, Cao2022GeometryAV, Yang_2021_LayoutTransformer,inoue2023flexdm,wang2023box_net} and diffusion models~\cite{ho2020denoising, hui2023unifying, inoue2023layoutdm, chai2023layoutdm,fengheng}, have been employed to learn the underlying patterns and structures of layouts from large datasets. These methods can be roughly classified into two types: content-agnostic and content-aware methods. Content-agnostic methods place elements on a blank canvas and thus have fewer constraints. Instead of directly synthesizing pixel-wise layout, LayoutGAN~\cite{Li2019LayoutGANGG} predicted the labels and geometric parameters of elements.
BLT~\cite{BLT2022ECCV} introduced a bidirectional transformer to generate all results simultaneously for faster speed and better controllability. 
LayoutDM~\cite{inoue2023layoutdm} devised a diffusion framework for categorical variables in discrete space.

However, unaware of the visual contents, content-agnostic methods fail to yield satisfactory poster layouts. Therefore, content-aware methods that predict layouts based on pre-generated images are receiving increasing attention in poster layout generation tasks. CGL-GAN~\cite{cgl-gan} and its subsequent work PDA-GAN~\cite{xu2023unsupervised} made attempts to narrow the domain gap between inpainted training images and clean testing images. DS-GAN~\cite{Hsu-2023-posterlayout} devised a CNN-LSTM framework to strike a balance between graphic and content-aware metrics. RADM~\cite{fengheng} first introduced the diffusion model in poster layout generation. It gradually converts a set of random boxes to a plausible layout. Although these methods produce reasonable poster layouts, they have strict requirements for image content. When image composition is complex, they are unable to find suitable places for elements. 
Unlike existing methods, our proposed PlanNet enables planning of the overall layout, allowing the RenderNet to reserve places for visual elements.

\subsection{Conditional Image Generation}
The conditional generation models aim to generate images and videos according to the given conditions~\cite{DiffFashion, model, Harmonization, Language-Guided, Semantic}. At the rendering stage of our poster generating system, the generator should consider the product image and the layout of visual components, which is related to the research of image-inpainting~\cite{ku2023staging,lama,jiahui,repaint,blended,controlnet,Rombach_2022_CVPR, mutual}, and layout-to-image~\cite{he2021context,chen2023training,couairon2023zero,zheng2023layoutdiffusion,yang2023law,xue2023freestyle,controlnet}.

Image-inpainting methods complete the masked area of an image based on the known area. LaMa~\cite{lama} proposed a Fast Fourier Convolution module to fill the image with higher receptive fields. Yu \etal~\cite{jiahui} proposed a contextual attention layer to extract features from remote regions. With the rise of the diffusion model, more and more researchers proposed image-inpainting methods based on the Denoising Diffusion Probabilistic Models~\cite{repaint, controlnet, blended, Rombach_2022_CVPR}. For example, Stable Diffusion (SD)~\cite{Rombach_2022_CVPR} used the U-Net~\cite{unet} module to predict the noise with the given mask and image. ControlNet~\cite{controlnet} regard the incomplete image as the condition and used an extra brunch to control the inpainting process. To generate the background of product poster, the masked area is set as the background of the poster and should be completed based on the known area of the product. For example, Ku \etal~\cite{ku2023staging} proposed a GAN-based method to produce the background of the product ads. They copied the background from the ads of a similar product and modified the image by the GAN-based inpainting method. Although these methods can generate realistic backgrounds, they may not be suitable for placing text due to the complexity of the patterns.

To strictly control the relative positions of the items in the images, many researches add the layout information as a restriction to generate images. 
Yang \etal ~\cite{yang2023law} divided each layout image into several patches and fused the different layout images according to the position of the patch. Xue \etal ~\cite{xue2023freestyle} proposed a cross-attention module that aligned the text description with layout information, which can generate objects according to the text at the corresponding position. 
Zhang \etal ~\cite{controlnet} controlled the generation process by incorporating layout image features into the original stable diffusion model. They introduced an additional encoder branch to introduce the extra control conditions to the pre-trained SD module. However, these methods generate the background without perceiving the product information, which leads to a mismatch between the style of the generated background and the product. Therefore, we integrate the product image into the layout-to-image generator, combining the advantages of image-inpainting and layout-to-image methods.
\section{Method}
\label{sec:method}
\subsection{Task Definition}

\begin{table}
  \centering
  \resizebox{0.7\linewidth}{!}{ 
  \begin{tabular}{ccccc}
    \toprule
    Tasks & Background & Text & Product &  Layout\\
    \midrule
    PLG & $\times$ & $\times$ & $\checkmark$ & $\checkmark$ \\
    PG & $\times$ & $\checkmark$ & $\checkmark$ & $\times$\\
    BG & $\checkmark$ & $\times$ & $\checkmark$ & $\times$ \\
    VTG & $\checkmark$ & $\checkmark$ & $\times$ & $\times$\\
    PPG & $\checkmark$ & $\checkmark$ & $\checkmark$ & $\checkmark$ \\
    \bottomrule
  \end{tabular}
  }
  \caption{Comparison of PPG with other tasks.}
  \label{tab-ppg}
\end{table}

Product Poster Generation (PPG) is defined as generating product poster images that contain given products and text content. Tasks similar to PPG include Poster Layout Generation (PLG)~\cite{cgl-gan,fengheng}, Poster Generation (PG)~\cite{ali-poster1}, Background Generation (BG)~\cite{ku2023staging,wang2023generate}, and Visual Text Generation (VTG)~\cite{yang2024glyphcontrol,chen2024textdiffuser, tuo2023anytext}. There are four distinct factors that make PPG more challenging compared to these tasks as outlined in Table~\ref{tab-ppg}: 1) \textit{Background}: The background image is generated automatically rather than manually designed. 2) \textit{Text}: Generated images need to include visual text with the given content. 3) \textit{Product}: The identity of the product shown in posters should be exactly consistent with the given product. 4) \textit{Layout}: The positions of products and texts are automatically generated rather than manually designed.

\subsection{Overall Framework}
\begin{figure*}
  \centering
    \includegraphics[width=\linewidth]{ 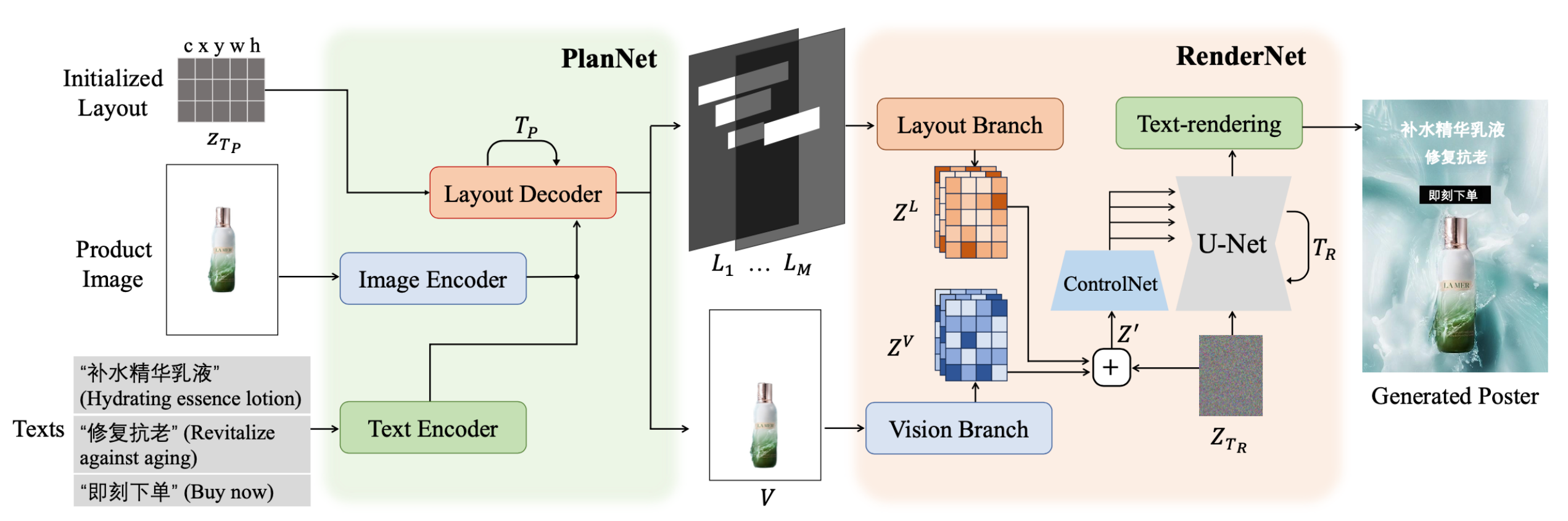}
    \caption{The framework of P\&R. It consists of a PlanNet and a RenderNet. The PlanNet aims to generate layouts based on the product image and texts, and the RenderNet aims to generate posters based on the product image and the generated layouts from PlanNet. The gray blocks contain texts and translations in the brackets. }
    \label{fig: framework}
\end{figure*}

The framework of P\&R is shown as Figure~\ref{fig: framework}, which consists of a PlanNet and a RenderNet. The PlanNet predicts the overall layout of the poster based on the given product image and texts, and then the RenderNet renders the predicted layout as a poster image. For the PlanNet, it first encodes the product image and texts with the image and  text encoders. Then we input the encoded features to a layout decoder based on discrete diffusion model to learn the relationship among layout elements, image features, and text features. After a $T_P$-step denoising process, the initialized layout is restored to the overall layout of the poster.
For the RenderNet, it utilizes the layout results generated by PlanNet with two branches. The first layout branch fuses the layouts of visual components into a unified layout embedding with a spatial fusion module, which can learn the spatial relationship among the different visual components. The second vision branch uses a series of convolution layers to extract the visual embeddings of the repositioned product, which contributes to generating a harmonious background with the product. 
Then extracted layout and visual embeddings are put into the ControlNet to guide the generation process of stable diffusion. After generating an image prepared for rendering texts, we use a text-rendering module to obtain the final product poster. 

\subsection{PlanNet}
A prior overall planning layout is beneficial for the downstream generation tasks. Although content-agnostic methods can predict the overall layout, they neglect the appearance of the product and the text content, making it difficult to achieve a good visual balance. To address this issue, we propose the PlanNet, which considers the product and text information simultaneously while predicting the overall layout.
The layout consists of a set of elements, which are represented by five attributes $c, x, y, w, h$, where $c$ is the category, $(x,y)$ is the center point, and $w$, $h$ are the width and height.
Each attribute has $K$ different values.
To handle discrete variables, the PlanNet is based on discrete diffusion models including diffusion and denoising processes.

\begin{figure}
  \centering
    \includegraphics[width=1.\linewidth]{ 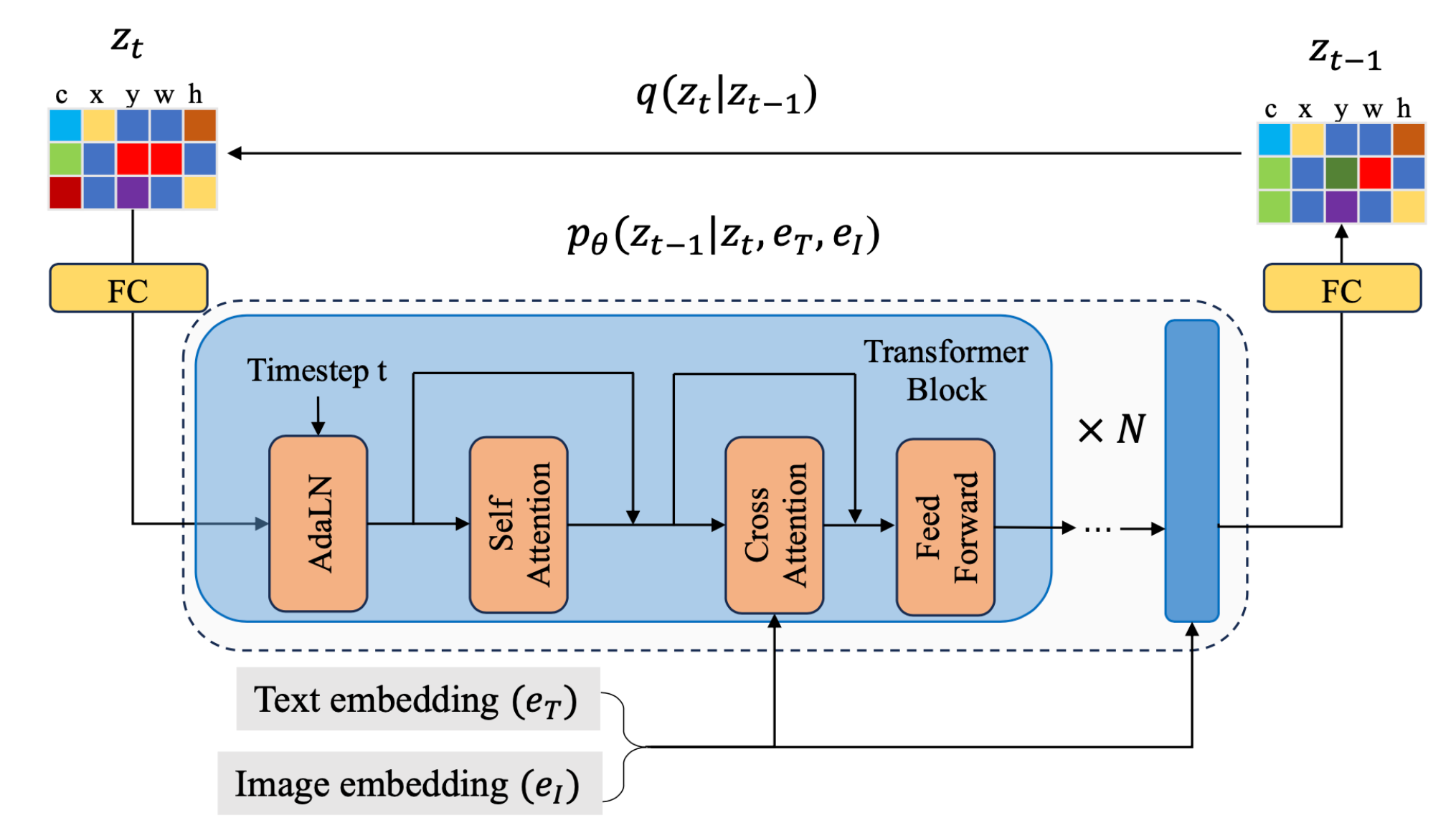}
    \caption{The architecture of the layout decoder.}
    \label{fig: PlanNet}
\end{figure}

In the diffusion process, we denote attribute at time $t$ as $z_t$,  $z_t \in \{1, 2, \cdots, K\}$.
The probability of transition from $z_{t-1}$ to $z_t$ is defined as $[Q_t]_{ij} = q(z_t=i | z_{t-1}=j)$, and the state transition matrix $\mathbf{Q_t}$ is defined as the following:
\begin{equation}
\mathbf{Q_t} = \begin{bmatrix}
\alpha_t & \beta_t & \beta_t & \hdots & 0\\
\beta_t & \alpha_t & \beta_t & \hdots & 0 \\
\beta_t & \beta_t & \alpha_t & \hdots & 0\\
\vdots  & \vdots  & \vdots  & \ddots  & \vdots\\
\gamma_t & \gamma_t & \gamma_t & \gamma_t & 1
\end{bmatrix},
\end{equation}
where $\alpha_t, \beta_t, \gamma_t$ are the probabilities to keep the same state, change to other states, and change to $[{\rm MASK}]$ state~\cite{gu2022vector} at timestep $t$, respectively.
Once the attribute changes to $[{\rm MASK}]$ state, it will not change to other states.
When $t$ is sufficiently large, all states will eventually fall into $[{\rm MASK}]$ state.
The forward process in matrix format can be written as:
\begin{equation}
    q(z_t|z_{t-1}) = \mathbf{v}^T(z_t)\mathbf{Q_t}\mathbf{v}(z_{t-1}),
\end{equation}
where $\mathbf{v}(z_t)$ is a column one-hot vector of $z_t$.
Due to the property of Markov chain $q(z_t|z_{t-1}, z_0) = q(z_t|z_{t-1})$, the probability of transition from $z_0$ to $z_t$ can be simplified as:
\begin{equation}
    q(z_t|z_0) =  \mathbf{v}^T(z_t)\mathbf{\mathbf{Q_t} \mathbf{Q_{t-1}} \dots \mathbf{Q_1}}\mathbf{v}(z_{0}).
\end{equation}
Conditioned on $z_0$, we can also calculate the posterior of diffusion process by:
\begin{equation}
q(z_{t-1}|z_{t}, z_0)  = \frac{q(z_{t}|z_{t-1}, z_0)q(z_{t-1}|z_0)}{q(z_{t} | z_0)}. \\
\end{equation}

In the denoising process, we use the trained PlanNet to convert initialized layout $z_{T_P}$ to $z_0$.
PlanNet consists of an image encoder, a text encoder, and a layout decoder.
To make the encoders better adapted to e-commerce scenarios,
we have collected over 30 million image-text pairs from the e-commerce platform, and finetune ALBEF~\cite{li2021albef} following its original training objectives, including image-text contrastive learning, masked language model, and image-text matching.
Besides, we add another training objective which aims to predict the category of product based on image and texts.
Following the configuration in ALBEF, the image encoder is a 12-layer visual transformer ViT-B/16~\cite{dosovitskiy2020vit}, and the text encoder is the first 6 layers of RoBERTa~\cite{Liu2019RoBERTaAR} pretrained with Chinese. 
However, it is worth noting that our method is not limited to Chinese.
By changing the pre-trained text encoder, other languages can also be supported.

Then the layout decoder uses the extracted text and image embeddings to estimate the reverse process from $z_t$ to $z_{t-1}$.
As shown in Figure~\ref{fig: PlanNet}, it is comprised of two fully connected (FC) layers and $N$ transformer blocks.
Firstly, $z_t$ is projected to element embedding $e_t^0$ with an FC layer.
After $N$ transformer blocks, the element embedding $e_t^{N-1}$ is decoded into $z_{t-1}$ with another FC layer. In each transformer block, the time step $t$ is combined with element embedding $e_t$ using an adaptive layer normalization (AdaLN) and a self-attention (SA) layer:

\begin{equation}
    h_t  = {\rm AdaLN}(e_t, t),
\end{equation}
\begin{equation}
    a_t  = h_t + {\rm SA}(h_t).
\end{equation}
Then the cross-attention (CA) is calculated using the result of self attention and the concatenation of text embedding $e_T$ and image embeddings $e_I$:
\begin{equation}
        e_t = {\rm FF}( a_t+ {\rm CA}(a_t, {\rm CAT}(e_T, e_I))),\\
\end{equation}
where ${\rm CAT}$ is concatenation and ${\rm FF}$ is the feed forward function.

\begin{figure}
  \centering  
    \includegraphics[width=1.\linewidth]{ 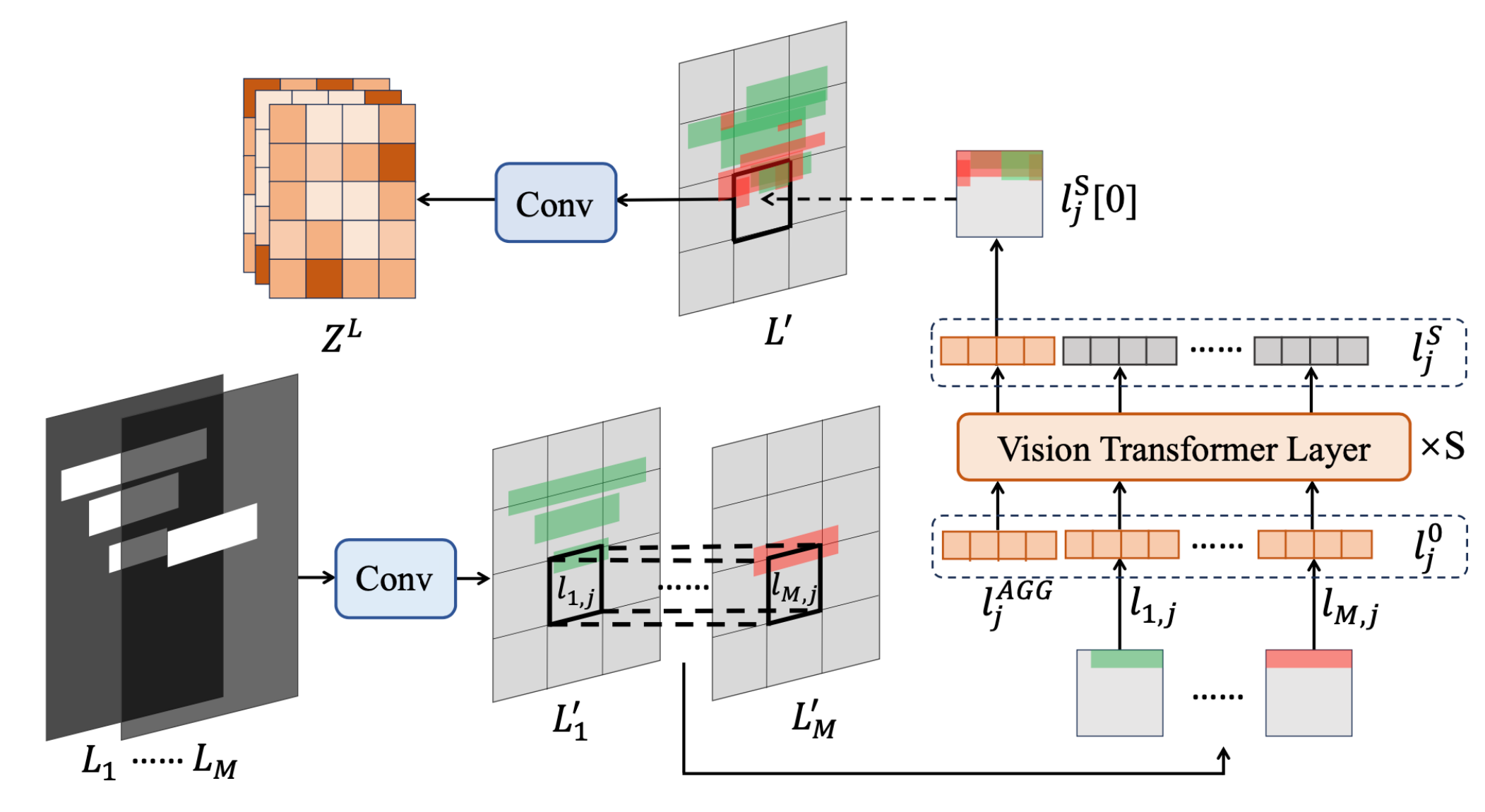}
    \caption{The architecture of the spatial fusion module.}
    \label{fig: sub}
\end{figure}

\subsection{RenderNet}
The background generated by image-inpainting methods may fail to place texts because of the complex patterns, therefore we put forward the RenderNet that incorporates layout information generated by PlanNet into the image-inpainting. 
RenderNet consists of a layout branch, a vision branch, Stable Diffusion (SD)~\cite{Rombach_2022_CVPR}, ControlNet~\cite{controlnet} and a text-rendering module. The layout branch aims to encode layouts of visual components. To better represent the spatial information of layouts, we transform the coordinate information from the PlanNet to masked layout images $\{L_m\}$, where $m=1,...,M$ and $M$ is the category number of visual components. For $L_m$, the areas of the $m$-th type of visual component are filled with 1 and the rest area is 0. 
To exploit the spatial relationship among $M$ layouts, we put forward a spatial fusion module shown as Figure~\ref{fig: sub}. 

Firstly, we use a three-layer convolution network to encode $\{L_m\}$ to latent representations $\{L'_m\}$, where $L'_m\in \mathbb{R}^{C\times W\times H}$. Then we fuse $\{L'_m\}$ into a unified layout representation $L'$. Specifically, $L'_m$ are divided into patches $\{l_{m,j}\}$ with the shape of $C\times P\times P$, where $j$ is the index of the patch and $j=1,...,W/P * H/P$. To obtain the $j$-th patch of $L'$, we fuse the $j$-th patch of different $L'_m$ as follows:
\begin{equation}
  l_j^0 = {\rm CAT}(l_j^{\rm AGG},l_{1,j}, ..., l_{M,j}),
  \label{eq1}
\end{equation}
where ${\rm CAT}$ is concatenation and $l_j^{\rm AGG}$ is an aggregation token added to the input. Then $l_j^0$ is input into a $S$-layer vision transformer:
\begin{equation}
  l_j^s = {\rm MHSA}(l_j^{s-1})+l_j^{s-1},
  \label{eq2}
\end{equation}
\begin{equation}
  l_j^s = {\rm FF}(l_j^s)+l_j^s,
  \label{eq3}
\end{equation}
where ${\rm MHSA}(\cdot)$ is the multi-head self-attention function, and ${\rm FF}(\cdot)$ is a linear network with layer normalization. We use the $l_j^{S}[0]$ as the $j$-th patch of $L'$. Finally, we apply a three-layer convolution network on $L'$ to get the layout embedding $Z^L$.

The vision branch aims to perceive the visual and spatial information of the product. We firstly zoom and move the product image based on the  coordinate information from PlanNet to get a repositioned product image $V$. Then we apply a six-layer convolution network on $V$ to extract the visual embedding $Z^V$. Finally, the visual and layout embeddings are added into ControlNet $\epsilon_\theta^c(\cdot)$ as control condition to guide SD.

At the training stage, we encode the prompt as $\tau$ by a text encoder $\tau_\theta$ of the pre-trained CLIP~\cite{clip}. At the $t$-th time-step, a random noise $\epsilon$ is added to the embedding of the training image $Z$ to produce the noised feature $Z_t$. Then, we use a U-Net $\epsilon_\theta(\cdot)$ to predict the added noise $\epsilon$. The loss function can be formulated as follows:
\begin{equation}
  \mathcal{L} = \mathbb{E}_{Z, \tau, t, \epsilon\sim \mathcal{N}(0,1)}[\|\epsilon - \epsilon_\theta(Z_t,t,\tau_\theta(\tau), \epsilon^c_\theta(Z'))\|_2^2],
  \label{eq5}
\end{equation}
where $Z' = Z_t + Z^L + Z^V.$

At the inference stage, the well-trained RenderNet predicts the noise added to an image for $T_R$ steps,
given a random noise $Z_{T_R}$, text embedding $\tau$, the control condition $\epsilon^c_\theta(Z')$. 
After removing the noise, we obtain the image $Z_0$ prepared for placing texts.  

\label{sec:dataset}
\begin{figure}
  \centering
    \includegraphics[width=\linewidth]{ 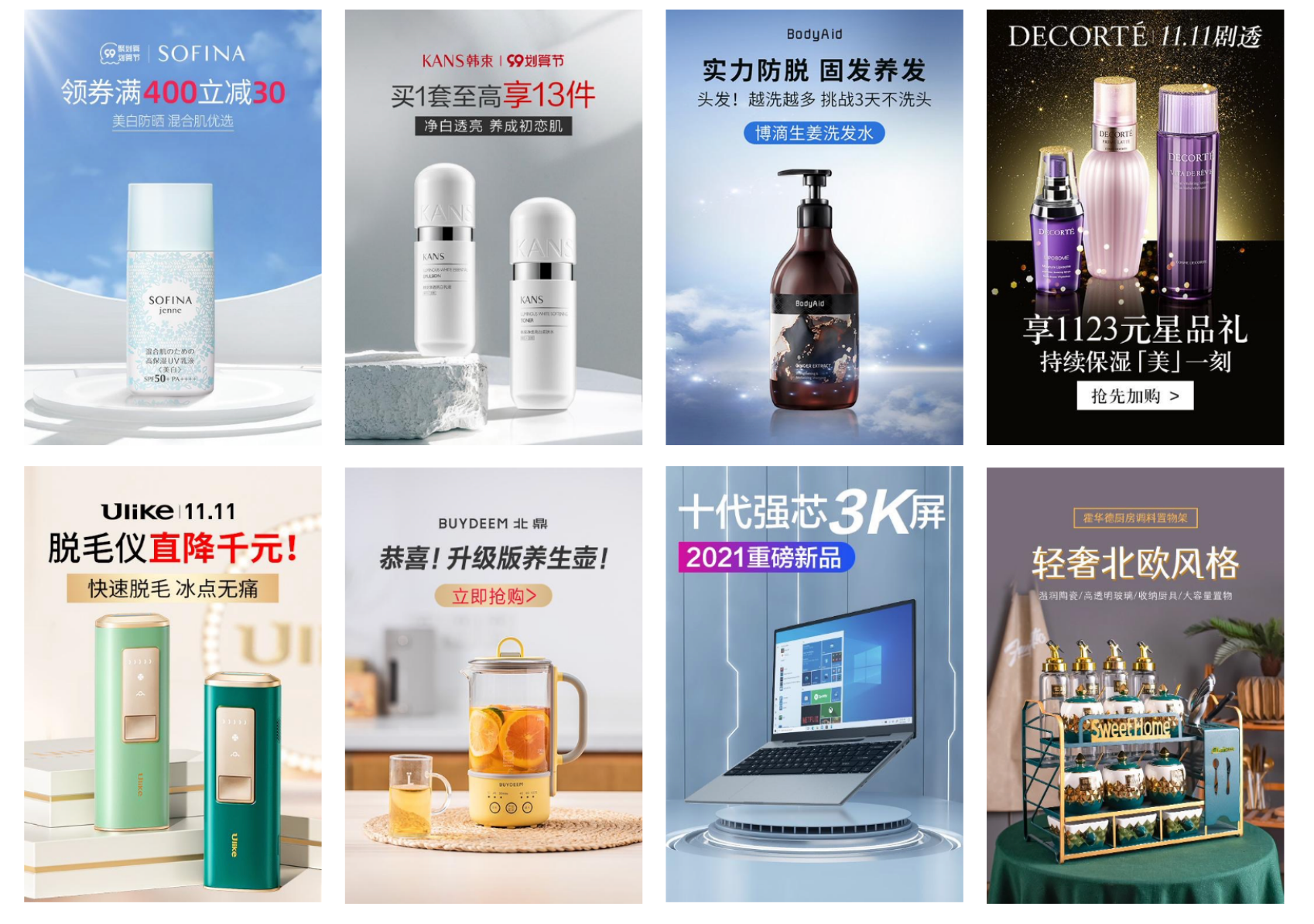}
    \caption{Some visual examples in PPG30k.}
    \label{fig: dataset_example}
\end{figure}

As it is insufficient to train a visual text generation model solely on the poster image dataset, we introduce an extra text-rendering module to render texts on the generated image heuristically. Specifically, assuming that the location of the i-th text element that needs to render  is ${Text}_i$. To determine the text color, we first filter out pixels with lower transparency within ${Text}_i$ of $Z_0$ to obtain a set of pixels that contribute more to visual perception. Then, we use the color with the highest proportion in the set as the representative color of ${Text}_i$. Finally, we offset the representative color in HSV space as the text color, and the offset value needs to ensure that the color tone before and after the offset is consistent, and the aesthetic matching is harmonious. To determine the text font, we collect over twenty commonly used font libraries in advance, and then randomly select one as the text font. The relevant code will be released to help reproduce the results. Although some pre-trained text generation and editing models~\cite{yang2024glyphcontrol,chen2024textdiffuser,tuo2023anytext} can also be used to render text in designated areas, we have empirically found that the generated text is prone to errors, especially for Chinese, and the heuristic rule-based text rendering module can accurately and clearly generate visual text. In the future, we will focus on improving the ability of visual text generation models and inheriting them into RenderNet.


\section{Dataset}

\begin{figure}
  \centering
    \includegraphics[width=.9\linewidth]{ 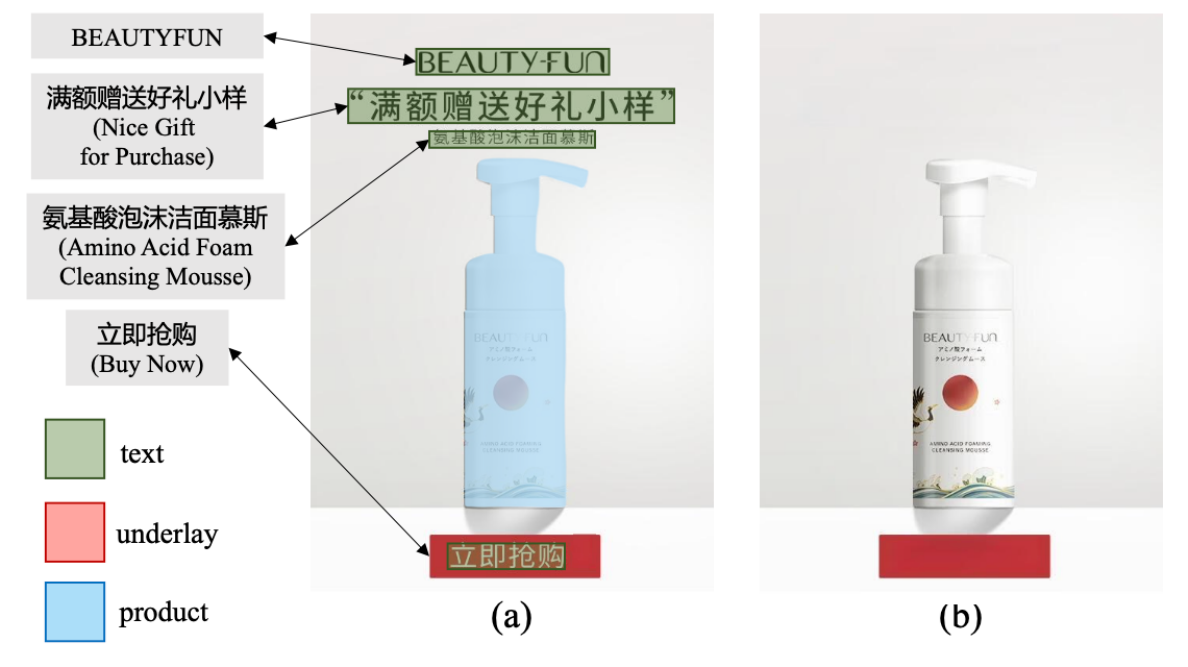}
    \caption{(a) Poster layout annotations. Product, texts and underlay are annotated. The texts in gray boxes are the translations of the texts on the posters. (b) The clean image. The texts on the poster are erased while underlays are retained.}
    \label{fig: dataset}
\end{figure}

To promote research in this field, we construct the first large-scale product poster generation dataset PPG30k. The images are collected from CGL-Dataset~\cite{cgl-gan}, where we remove posters with portraits to avoid portrait rights disputes, and posters with unsightly backgrounds to improve generation quality.
The selected images are with pleasing backgrounds and can effectively transmit messages, as shown in Figure~\ref{fig: dataset_example}.
Eventually, there are 32724 training images and 1426 testing images in PPG30k, and the image size is $513\times750$.

\begin{figure*}
  \centering  
    \includegraphics[width=\linewidth]{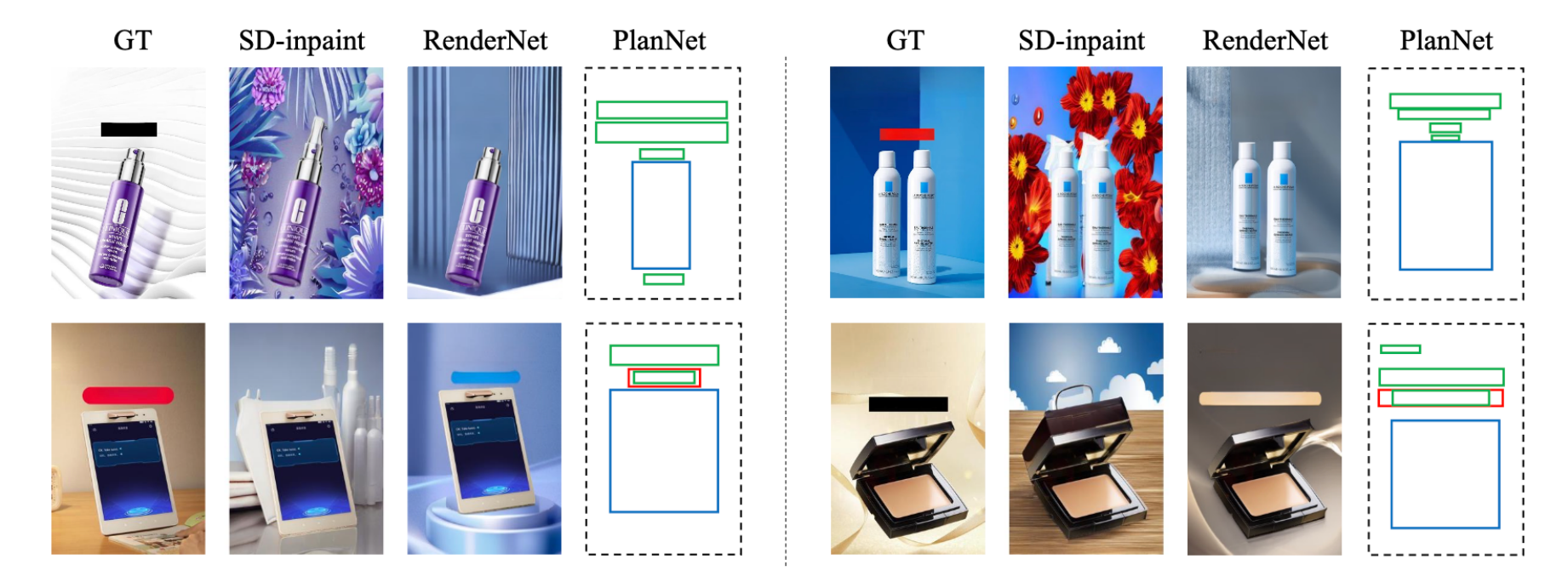}
    \caption{Qualitative comparison with SD-inpaint. ``GT'' refers to the ground truth. The green, red and blue boxes respectively represent texts, underlays and the product (the same below).}
    \label{fig: pipeline}
\end{figure*}

As illustrated in Figure~\ref{fig: dataset} (a), there are three types of elements including texts, underlays, and products.
The geometric attributes $\{x, y, w, h\}$ and categories $c$ are annotated.
Besides, as texts play an important role in conveying messages in posters, the corresponding textual contents are also recorded.  
To get the precise product appearance, products are annotated with masks.
When training the PlanNet, we use the smallest rectangular area that can surround the product as its location.
The PlanNet takes the textual contents and product images as input, and the location and categories are the ground truth. For convenience, we will release extracted image and text embeddings.
When training the RenderNet, we convert the location of texts and underlays to masks.
As shown in Figure~\ref{fig: dataset} (b), the RenderNet takes the masks and product images as input, and the clean images are used as ground truth where texts are erased.

\section{Experiments}
\label{sec:exp}

\subsection{Implementation Details}
Since RenderNet requires precise control, and there is noise in the layouts predicted by PlanNet, which is harmful to network training. Therefore, we train PlanNet and RenderNet separately and use ground-truth layouts when training RenderNet.

\begin{table}
  \centering
  
  \begin{tabular}{ccccc}
    \toprule
    Method & FID-image $\downarrow$ & CLIP-score $\uparrow$ & BQ $\uparrow$ & LR $\uparrow$\\
    \midrule
    SD-ipt~\cite{Rombach_2022_CVPR} & 47.30 & 87.60 & 2.20& 2.06 \\
    CN-ipt~\cite{controlnet} & 67.19 & 79.54 & 1.34& 1.79\\
    LaMa~\cite{lama} & 50.59 & 87.89 &2.82 & 2.88\\
    P\&R & \textbf{35.23} & \textbf{91.80} & \textbf{3.66}& \textbf{3.24} \\
    \bottomrule
  \end{tabular}
  \caption{Comparison results with SOTA poster generators, which consist of the image-inpainting method and CGL-GAN~\cite{cgl-gan}. ``SD-ipt'' and ``CN-ipt'' refer to the inpainting version of stable diffusion and ControlNet.}
  \label{tab1}
\end{table}

For the PlanNet, there are 4 transformer blocks with 8 attention heads, 512 embedding dimensions, and 2048 hidden dimensions, and we set the dropout rate 0.1.
The diffusion timesteps $T_P$ is 100. 
We train the PlanNet for 100 epochs using AdamW optimizer \cite{loshchilov2017adamw} with a learning rate of $5.0 \times 10^{-4}$.
The training time is 12 hours on a single P40 GPU.
For the RenderNet, it takes 1 day on 24 A100 GPUs to train  for 500 epochs with the same optimizer as ControlNet~\cite{controlnet}.  The hidden feature map of the spatial fusion module satisfies $C=320, W=128, H=192$. The layer number $S$ is 2, the inference step $T_{R}$ is 50, and the patch size $P$ is 8. 
\subsection{Evaluation Metrics}

\begin{table}
\centering
 \begin{tabular}{c c c c c c c}
    \toprule
    Method & FID-layout $\downarrow$ & Max IoU $\uparrow$ &  FID-image $\downarrow$ & CLIP-score $\uparrow$ & ${\rm R_{pass}}$ $\uparrow$ & ${\rm R_{best}}$ $\uparrow$\\
        \midrule
    LayoutDM~\cite{inoue2023layoutdm}    & 15.60 & 0.51 & 38.97 & 89.89 & 53.95\%  & 20.65\%\\
    FlexDM~\cite{inoue2023flexdm}      & 107.33& 0.21 & 52.88 & 81.20 &58.40\%&24.60\% \\
    BoxNet~\cite{wang2023box_net}    & 93.54 & 0.35 & 43.60 & 88.28 &39.85\% &12.35\%\\
    P\&R        & \textbf{14.50} & \textbf{0.58} & \textbf{35.23} & \textbf{91.80} & \textbf{70.85\%}& \textbf{42.40\%}\\
    \bottomrule
  \end{tabular}
  \caption{Comparison results with SOTA layout generators. }
  \label{tab2}
\end{table}

\textbf{Numerical Metrics.} To evaluate the PlanNet, Frechet Inception Distance for layout (FID-layout)~\cite{inoue2023layoutdm} is used to measure the similarity between the generated and ground truth layouts. Maximum Intersection over Union (Max IoU)~\cite{kikuchi2021miou} is used to compute the average IoU between the optimal matching between the generated boxes and the ground truth with the same category. 
To evaluate the RenderNet, Frechet Inception Distance for image (FID-image) is used to measure the distance between the distributions of generated and real images. CLIP-score~\cite{clip} is used to calculate the cosine similarity between the features of generated and real images, which are extracted by the image encoder of a pre-trained CLIP model. 

\noindent
\textbf{User Study.} To evaluate the quality of the poster, we conduct a user study on 20 experienced advertising practitioners. In this study, we use 100 groups of images, where each group contains 4 posters generated by different methods. All these results in each group are presented side-by-side and in a random order. Practitioners are asked to rank them from the perspectives of background quality (BQ) and layout rationality (LR), where 1 is the worst and 4 is the best. When ranking the BQ, the practitioners focus on the suitability of serving as the product poster and the harmony between the background and the product. When ranking the LR, the practitioners consider the overlap of different components and the visual balance of components on the poster.

To avoid the influence of font on the ranking results of BQ and LR, there is no text on the poster, and all the visual components are represented by boxes.
To evaluate the performances of poster layout generators, we use 100 groups of layouts, where each group contains 4 randomly ordered layouts generated by different methods. Practitioners are asked to judge whether the layout result is qualified (${\rm R_{pass}}$) and select the best layout (${\rm R_{best}}$). When evaluating the two metrics, practitioners consider the overlap of different components and, rationality of visual components. 

\subsection{Quantitative Analysis}

\begin{table}
\centering
 \begin{tabular}{c c c c}
    \toprule
    Method & $R_{com} \downarrow$ & $R_{sub} \downarrow$ &  $R_{shm}$ $\downarrow$ \\
        \midrule
    LayoutVTN~\cite{arroyo2021variational}      & 41.77 & 1.32 & 22.21  \\
    LayoutTransformer~\cite{Yang_2021_LayoutTransformer}   & 40.92 & 1.31 & 21.08 \\
    LayoutDM~\cite{inoue2023layoutdm}  & 40.91 & 1.24 & 18.42 \\
    PlanNet  & \textbf{39.31} & \textbf{1.02} & \textbf{15.39} \\
    \bottomrule
  \end{tabular}
  \caption{Layout generation results on CGL-Dataset.}
  \label{cgl}
\end{table}

\begin{table}
\centering
 \begin{tabular}{ccc}
    \toprule
    Method & FID-image $\downarrow$ & CLIP-score $\uparrow$ \\
    \midrule
    CN-lyt~\cite{controlnet} & 44.31 & 69.92 \\
    Freestyle~\cite{xue2023freestyle} & 68.68 & 60.11\\
    P\&R & \textbf{33.55} & \textbf{91.46}\\
    \bottomrule
  \end{tabular}
  \caption{Comparison results with SOTA conditional image generators. ``CN-lyt'' refers to the layout-to-image version of ControlNet.}
  \label{tab3}
\end{table}

\begin{table}
\centering
 \begin{tabular}{cccc}
    \toprule
  Text & Image & FID-layout $\downarrow$ & Max IoU $\uparrow$ \\
        \midrule
        \checkmark &\checkmark   & \textbf{14.50} & \textbf{0.58} \\
     \checkmark &                & 17.15 & 0.55  \\
    &\checkmark                 & 18.42 & 0.52  \\
    &                           & 21.05 & 0.51 \\
    \bottomrule
  \end{tabular}
  \caption{Ablation study of image and text contents. }
  \label{tab4}
\end{table}

\begin{figure*}
  \centering 
    \includegraphics[width=.95\linewidth]{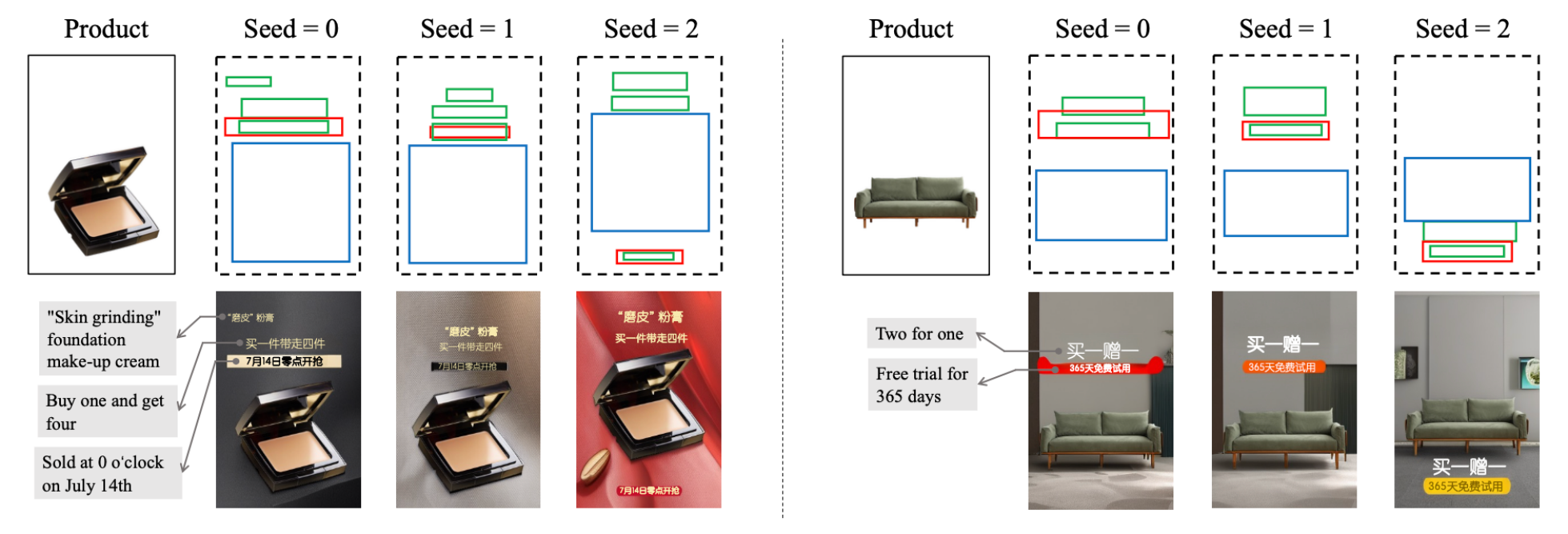}
    \caption{Generated posters under different random seeds. The texts in gray boxes are the translations of the texts on the posters.}
    \label{fig: diversity}
\end{figure*}

\begin{figure}
  \centering
    \includegraphics[width=.8\linewidth]{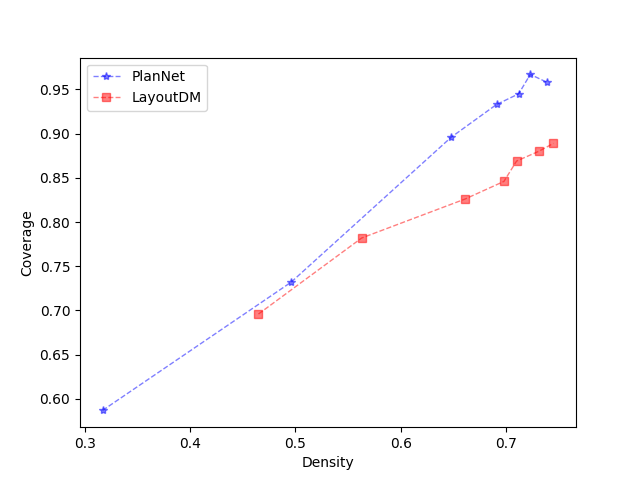}
    \caption{Density-coverage trade-off of different methods.}
    \label{con_den}
\end{figure}

\noindent
\textbf{Comparison with SOTA Poster Generators.}
As mentioned above, combining image-inpainting and poster layout generation methods can be seen as a naive solution to PPG. To verify the superiority of P\&R, we combine three pre-trained image-inpainting models with a SOTA layout generator CGL-GAN~\cite{cgl-gan} and compare their generated posters with ours. 

As Table~\ref{tab1} shows, our method performs the best on FID-image and CLIP-score, which indicates that the style of background images generated by our RenderNet is closer to that of product posters. We also calculate the average BQ and LR. The results show that the images generated by our method are more suitable to serve as product posters, and our generated layouts achieve a better visual balance. 
We take the SD-inpaint method as an example to compare our method with SOTA poster generators qualitatively. As Figure~\ref{fig: pipeline} shows, although the pre-trained SD-inpaint model can generate realistic background images, the content of the background images is too complex to have sufficient space for visual components, which results in a high failure rate of layout generation. Our method can generate an appealing background with the superiority of transmitting product information. 

\begin{figure}
  \centering
    \includegraphics[width=1.0\linewidth]{ 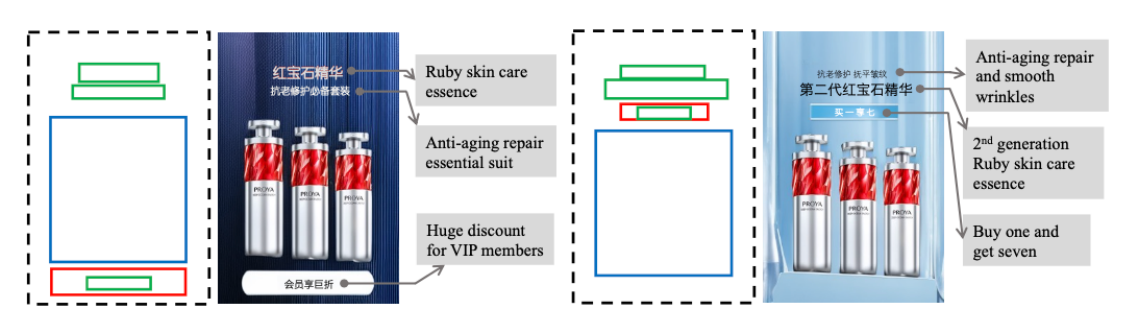}
    \caption{Generated results under different texts contents. The texts in gray boxes are the translations of the texts on the posters.}
    \label{text_ratio}
\end{figure}

\noindent 
\textbf{Comparison with SOTA Layout Generators.}
To verify the advantages of our PlanNet, we compare the PlanNet with SOTA layout generators, including LayoutDM~\cite{inoue2023layoutdm}, FlexDM~\cite{inoue2023flexdm} and BoxNet~\cite{wang2023box_net}.
Based on their official codes, we train these models on PPG30k.
The comparison result is shown in Table~\ref{tab2}.
Because of making fuller use of text and image features, our method achieves the best on FID-layout and Max IoU metrics.
Additionally, to verify the impact of layout on background generation, we use the RenderNet to generate background based on the layout results of four methods.
The images generated from the results of the PlanNet achieve the best on FID-image and CLIP-score metrics, which proves that our PlanNet is more beneficial for downstream image generation.
Meanwhile, the result of the user study indicates that our PlanNet arranges various elements with better composition. As shown in Table~\ref{cgl}, we also compare the performance on public CGL-Dataset, which further proves the superiority of PlanNet.
\begin{figure}
  \centering
    \includegraphics[width=1.\linewidth]{ 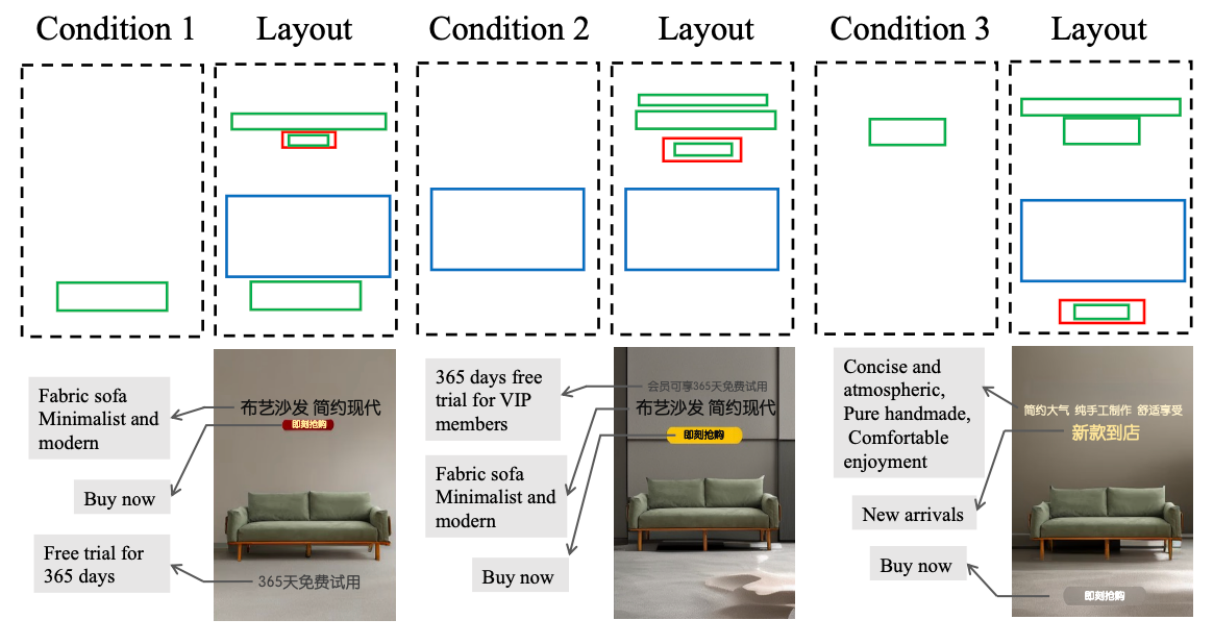}
    \caption{Generated results under different layout conditions. The texts in gray boxes are the translations of the texts on the posters.}
    \label{cond}
\end{figure}
\noindent 
\textbf{Comparison with SOTA Conditional Generators.}
Layout-to-image methods ignore the product content when generating posters, resulting in the generated background may not match the style of the product. Our RenderNet combines the advantages of both image-inpainting and layout-to-image methods, which can give attention to both layouts and the product.
To verify this, we finetune two state-of-the-art layout-to-image methods based on the ground truth layouts. Our P\&R method takes product images as additional inputs, which makes the generated background more suitable for specific products. Table~\ref{tab3} indicates that backgrounds generated by our method match the product more harmoniously.

\noindent 
\textbf{Effects of Image and Text Contents.}
To verify the influence of image and text contents on the layout, we remove the image encoder, text encoder, and both of them to train three variants of the PlanNet on PPG30k.
As Table~\ref{tab4} shows, removing image, text contents results in 2.65, 3.92 reduction on FID-layout and 0.03, 0.06 reduction on Max IoU, respectively. 
Removing both results in further performance degradation, which proves the significance of considering image and text contents in poster layout generation.



\subsection{Qualitative Analysis}

\noindent 
\textbf{Diversity.}
To show the diversity of P\&R, we set different random seeds to generate posters. 
As shown in Figure~\ref{fig: diversity}, the leftmost column is the input product and texts.
In the rest part, the first row is the different layouts and the second row is the posters generated based on the layouts above. 
This shows that our method can generate a variety of reasonable layouts and appealing posters. We also plot the density and coverage metrics~\cite{inoue2023layoutdm} of PlanNet and LayoutDM in Figure~\ref{con_den}, which proves that our method achieves a better trade-off between the diversity and fidelity.

   
\noindent 
\textbf{Different Text Contents.}
To verify that our method can effectively understand text contents, we keep the random seed unchanged and use different text contents as input. 
The result is shown in Figure~\ref{text_ratio}, which proves that our method can produce appropriate layouts and posters based on different text contents.

\noindent 
\textbf{Partial Layout Condition. }
Since our method supports that pre-defined conditions can be injected into the element embedding,
users can pre-set the position, size, and category of the elements according to their own preferences.
In Figure~\ref{cond}, we show the results under three commonly used layout conditions.
From left to right, we set a text at the bottom, a product in the middle, and a text at the top, respectively.
The generated results show that our method performs well in completing the entire layouts and posters based on partial layout conditions.

\section{Conclusion}

In this paper, we propose a novel Product Poster Generation (PPG) task and a corresponding product poster generation framework with diffusion models named P\&R, which can generate product posters with well-designed layouts and visually appealing backgrounds. In P\&R, a PlanNet is firstly used to plan the overall layout, which merges the product image and text contents. This facilitates the creation of a well-organized layout. Then a RenderNet employs a spatial fusion module to incorporate the layouts generated by PlanNet, and encodes the appearance of the product to ensure a harmonious blend between the generated background and the product image. To promote research in this field, we construct the PPG30k dataset for PPG task. Extensive experiments on PPG30k demonstrate that P\&R surpasses state-of-the-art methods in terms of layout prediction and poster quality. In the future, we will focus on integrating the visual text generation model into a unified framework.

\vfill


\bibliography{ref}

\end{document}